\documentclass[conference]{IEEEtran}
\IEEEoverridecommandlockouts
\usepackage[utf8]{inputenc}
\usepackage{graphicx}
\usepackage{amsmath, amssymb, amsfonts}
\usepackage{booktabs}
\usepackage{tabularx}
\usepackage{float}
\usepackage{multirow}
\usepackage{caption}
\usepackage{subcaption}
\usepackage{enumitem}
\usepackage{xcolor}
\usepackage{tikz}
\usetikzlibrary{positioning}
\usepackage{hyperref}
\usepackage{comment}
\usepackage{todonotes}
\usepackage[most]{tcolorbox}
\hypersetup{colorlinks=true, linkcolor=blue, citecolor=blue, urlcolor=blue}

\title{Behavioral Heterogeneity as Quantum-Inspired Representation}

\author{%
  \parbox{\textwidth}{\centering
    Mohammad Elayan, Wissam Kontar$^{\ast}$ %
  }%
  \thanks{University of Nebraska--Lincoln, Lincoln, NE, USA.
  {\tt\small melayan2@nebraska.edu, wkontar2@nebraska.edu}}%
}    

\begin{document}

\maketitle

\begin{abstract} 

Driver heterogeneity is often reduced to labels or discrete regimes, compressing what is inherently dynamic into static categories. We introduce quantum-inspired representation that models each driver as an evolving latent state, presented as a density matrix with structured mathematical properties. Behavioral observations are embedded via non-linear Random Fourier Features, while state evolution blends temporal persistence of behavior with context-dependent profile activation. We evaluate our approach on empirical driving data, Third Generation Simulation Data (TGSIM), showing how driving profiles are extracted and analyzed. \emph{Full reproducibility is supported via our open-source codebase on \href{https://github.com/wissamkontar/Behavioral-Heterogeneity-as-Quantum-Inspired-Representation}{GitHub}.}

\end{abstract}


\section{Introduction}

Behavioral heterogeneity is a fundamental driver of performance and instability in systems shaped by agent interactions, and in transportation systems it is especially acute. The enduring problem of congestion traces directly to the inherent variability of human driving behavior. Today, emerging driving technologies, including autonomous vehicles (AVs) and advanced driver assistance systems (ADAS), introduce a new dimension of heterogeneity. Mixed traffic is no longer simply a spatial coexistence of AVs and human driven vehicles (HDVs); it is an environment saturated with conflict arising from variation in human behavior, machine level design philosophy, and human machine interaction.

Historically, modeling frameworks have treated heterogeneity as an ancillary input rather than an intrinsic property of the traffic agent. Physics based models typically rely on stochastic formulations to capture behavioral variation \cite{laval2010mechanism, chen2012behavioral}, while end to end data driven approaches often adopt a one model fits all strategy, collapsing learning to a single representation of the average observed heterogeneity \cite{marcano2020review,zhou2024implications}. In both cases, heterogeneity becomes visible only to the extent it is explicitly encoded in the model design or sufficiently represented in the training data. Even when heterogeneity is modeled, it is often compressed into a semantic taxonomy. Drivers are labeled as ``aggressive” or ``timid.” This can support simplified analysis, but it misses the reality that human behavior is dynamic. Humans do not occupy static states; they transition between behavioral modes as context and internal state evolve. The value is not in the label itself, but in the logic governing transitions. By collapsing behavior into deterministic categories, current models discard the information contained in the transition process.

This gap becomes critical as we move toward agentic level systems, where agents learn from data generated by their own interactions within the environment. When those interactions become the primary training signal, we must understand how behavioral states form and how they change. Without the logic of the transition, we risk training agents on misinterpreted data; with it, we gain the ability to intentionally train toward specific, desired behaviors.

In this work, we move beyond labeling to enable free form discovery that lets the \emph{data speak for itself}. We propose a quantum inspired framework in which behavioral taxonomy is not a static label, but a density matrix representation. The structure of these density matrices allows us to capture latent behavioral states, their interactions, and their continuous evolution over time. The result is a model agnostic representation that preserves the richness of behavioral transitions and treats heterogeneity as an evolving state defined by contextual sensitivity and temporal persistence.

\subsection{Relevant work}

Heterogeneity in driving behavior has traditionally been addressed through extensions of classical physics based car following models. Variants of Newell’s model \cite{newell2002simplified}, including the L–L \cite{laval2010mechanism} and AB models \cite{chen2012behavioral}, introduce driver specific parameters to represent behavioral variability. While these models offer structural insight, their representation of heterogeneity is constrained by predefined parametric forms, limiting variability to what is explicitly encoded in the model.

The rise of large scale trajectory datasets has shifted attention toward data driven and stochastic approaches that infer behavioral structure directly from observations. Recent work blends multiple car following formulations probabilistically \cite{jiang2024generic}, conditions parameter inference on recent trajectory history to capture latent intent \cite{zhang2024calibrating}, or uses regime switching extensions that allow transitions among context dependent parameterizations \cite{zhang2025markov, zhang2025context}. Although these approaches improve empirical fit and capture dynamic variability, heterogeneity still enters through mixture weights or regime assignments, compressing behavioral diversity into aggregate representations rather than a continuously evolving latent structure.

Recent developments in quantum cognition and quantum inspired machine learning offer an alternative for representing complex, context dependent variability without reducing agents to fixed categories. Hilbert space formulations model cognitive states as superpositions that evolve under contextual influence, while density matrices provide a consistent representation of mixed states and uncertainty \cite{Andrei_Khrennikov_2020, Jerome_R_Busemeyer_2024}. In machine learning, density operator formulations with Random Fourier Features support flexible distributional modeling under quadratic measurement rules consistent with the Born rule \cite{Gonzalez2022DensityMatricesRFF}. Motivated by these properties, we adopt a density matrix framework to model driving behavior as a context modulated state that evolves over time, preserving behavioral richness while remaining statistically tractable.

%This study proposes a quantum-inspired state modeling framework to represent and analyze driver behavior using high-resolution trajectory data. Driver behavior is modeled as a time-evolving latent state represented by a density matrix \cite{Gonzalez2022DensityMatricesRFF}, allowing complex interactions among behavioral variables to be captured while preserving temporal persistence. Contextual factors modulate the influence of latent behavioral profiles, and observed behavior continuously updates the state. The framework is designed for trajectory datasets such as TGSIM and supports interpretation in terms of behavioral adaptation, inertia, and contextual sensitivity.

\section{Methods}

%\todo[inline]{The methods sections is segmented into lots of subsections, for no clear purpose. I think we should incorporate sections together as much as possible to make the paper flow better}

This section presents the proposed modeling framework. Before introducing the technical components, we briefly clarify the structural principles that qualify the method as quantum-inspired in terms of density-matrix-based learning.

\begin{figure}[ht!]
\centering
\includegraphics[width=0.98\linewidth]{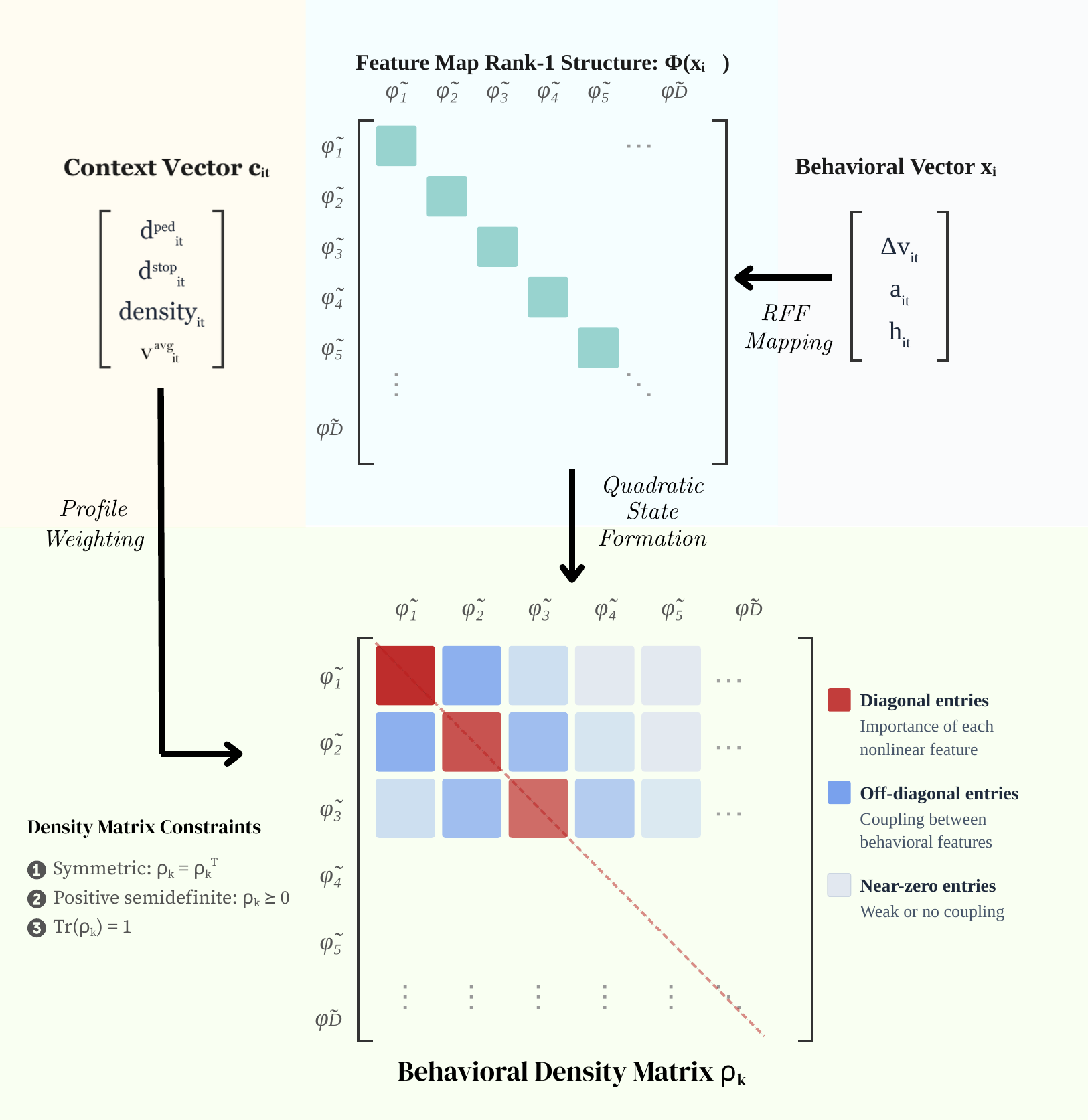}
\caption{Schematic of the proposed framework. Behavioral variables are mapped via RFF into a normalized feature space, where quadratic state formation yields density-matrix profiles. Context variables modulate profile weighting, and density-matrix constraints ensure valid state representation.}
\label{fig:dm_illustration}
\end{figure}

\begin{tcolorbox}[colback=yellow!6,
                  colframe=black!60,
                  boxrule=0.5pt,
                  arc=2pt]

\textbf{Quantum-Inspired Modeling Criteria.}
A learning framework is considered quantum-inspired if it satisfies three structural conditions:
(i) data are encoded as normalized states in a Hilbert space,
(ii) predictions are computed using quadratic measurement rules consistent with the Born rule, and
(iii) uncertainty is represented using valid density matrices that are symmetric, positive semidefinite, and trace-normalized \cite{Gonzalez2022DensityMatricesRFF}.

The proposed framework (Figure~\ref{fig:dm_illustration}) satisfies these conditions through its normalized Random Fourier Feature (RFF) representation, density-matrix behavioral profiles, and quadratic likelihood evaluation defined in the subsections below.

\end{tcolorbox}

%\todo[inline]{Figure 1 is never referenced in text}
%\todo[inline]{We need to be consistent with the naming. Sometimes we say behavioral vector, sometime physical state, sometimes feature vector. Let choose one and stick to it}

The remainder of this section formalizes each modeling component, beginning with features representation, followed by density-matrix construction, state evolution, and estimation.

\subsection{Data Representation and Nonlinear Feature Mapping}

Let $i$ index drivers (vehicles) and $t$ index discrete time steps. Each observation $(i,t)$ corresponds to one trajectory record, sampled at a fixed temporal resolution (e.g., 0.1~s).

Let \(x_{it} \in \mathbb{R}^d\) denote the behavioral vector describing how driver $i$ operates the vehicle at time $t$. The behavioral vector consists of continuous kinematic and interaction measures derived directly from trajectories. Our behavioral representation here is constructed as:

\[
x_{it} = 
\begin{bmatrix}
\Delta v_{it} & a_{it} & h_{it}
\end{bmatrix}^{\!\top}, 
\quad d = 3.
\]

where $\Delta v_{it}$ is relative speed with respect to the immediate leader, $a_{it}$ is signed scalar acceleration, and $h_{it}$ is headway distance to the leader. These variables form a minimal basis for longitudinal interaction, as classical car-following behavior is governed by relative speed, acceleration response, and spacing. While the observable state is compact ($d=3$), we do not assume the driver's response is simple or linear. In practice, drivers often behave nonlinearly. For instance, the same speed can feel ``safe" at one headway and ``dangerous" at another, or small changes in headway may matter when the gap is short but little when it is large. Explicitly modeling these interactions would require many higher-order terms (e.g., $\Delta v^2$), increasing model complexity and computational burden.

RFF provides a practical way to get that non-linear representation without changing the physical meaning of the behavioral state ($x_{it}$). The key idea here is that instead of manually inventing non-linear features, we automatically create a large set of non-linear alternative to the same three inputs in the behavioral state ($x_{it}$). RFF does this by passing the behavioral state through randomly generated wave functions (shown in Equation~\ref{eq:RFFcosine}). In effect, the original 3-dimensional behavioral vector ($x_{it}$) is re-expressed as an enriched behavioral vector of dimension \(D\), where each entry is a different transformation of the same underlying driving state.

Specifically, behavioral vectors are mapped into a higher-dimensional nonlinear RFF representation. We define a mapping \(\phi : \mathbb{R}^d \rightarrow \mathbb{R}^D\) with components:

\begin{equation}\label{eq:RFFcosine}
\phi_j(x) = \cos(w_j^\top x + b_j), \quad j = 1,\dots,D
\end{equation}

where $w_j \sim \mathcal{N}(0,\sigma^{-2}I_d)$, and $b_j \sim \mathrm{Uniform}(0,2\pi)$. This construction approximates a Gaussian radial basis function kernel and enables nonlinear interactions to be represented linearly in the RFF feature space. The parameters $\{w_j,b_j\}$ are sampled once at model initialization and maintained for all drivers at all time steps. Thus, the nonlinear RFF representation defines a global transformation of the behavioral vector.

The nonlinear RFF representation is normalized to ensure that quadratic forms defined on the feature space admit a probabilistic interpretation. Normalization is then: 

\begin{equation}
\tilde{\phi}(x_{it}) = \frac{\phi(x_{it})}{\|\phi(x_{it})\|}
\end{equation}

\subsection{Behavioral Profiles as Density Matrices}
\label{subsec:density_matrices}

%\todo[inline]{explain in layman terms what is a behavioral profile, give examples. For instance, these are congruent with the common taxonomy of aggressive, timid, neutral, etc.. however here represented as a matrix density function that itself is a taxomy yet not limited to a discrete choice of behavior, it holds higher level of information thanks to the RFF and the quantum-like ....}

The nonlinear RFF representation allows recurring patterns of driving behavior to be identified directly from trajectory data. These patterns may resemble familiar behavioral types such as cautious or assertive driving, but they are not predefined categories. Instead, they are learned from how drivers operate in the behavioral vector $(\Delta v, a, h)$.

We assume $K$ latent behavioral profiles. Each profile represents a structured pattern in the nonlinear RFF representation of the behavioral vector. Profile $k$ is represented by a density matrix $\rho_k \in \mathbb{R}^{D \times D}$ satisfying

\begin{equation}
\rho_k = \rho_k^\top,\quad \rho_k \succeq 0,\quad \mathrm{Tr}(\rho_k) = 1.
\end{equation}

Each $\rho_k$ is defined over the nonlinear RFF representation $\tilde{\phi}(x)$. In practical terms, the original behavioral vector $(\Delta v, a, h)$ is first mapped into its nonlinear RFF representation. The matrix $\rho_k$ then captures how these nonlinear RFF components interact. 

As shown in Figure~\ref{fig:dm_illustration}, diagonal entries reflect the relative importance of individual nonlinear RFF representation components, while off-diagonal entries encode their interactions. Unlike assigning a driver to a discrete type, each density matrix represents a distribution over behavioral modes, allowing mixed or transitional patterns to manifest in one profile.

\subsection{Contextual Activation and State Evolution}

%\todo[inline]{It is not clear why context matters in our formulation, we need to offer details, intuition, reasoning. It reads like a collection of equations rather than a storyline. Also discuss what happens if we do not do this step}

Driving behavior does not depend only on the internal car-following state depicted by $(\Delta v, a, h)$. 
It also depends on surrounding traffic and infrastructure conditions. For example, a driver may maintain similar relative speed and headway yet respond with more reactive acceleration when traffic density is high or when approaching a pedestrian zone. The role of context in our framework is therefore to explain when a driver is likely to express one behavioral profile versus another.

Let \(c_{it} \in \mathbb{R}^q\) denote the context vector. Context does not modify the nonlinear RFF representation of the behavioral vector directly. Instead, it determines how strongly each population-level behavioral profile $\rho_k$ is activated at time $t$ to form the driver-specific state. We define:

\[
c_{it} =
\begin{bmatrix}
d^{\text{ped}}_{it} & d^{\text{stop}}_{it} & \mathrm{density}_{it} & v^{\text{avg}}_{it}
\end{bmatrix}^{\!\top},
\quad q = 4.
\]

The first two variables ($d^{\text{ped}}_{it}$ and $d^{\text{stop}}_{it}$) are primarily relevant in urban traffic, capturing proximity to pedestrians and intersection control. The latter two ($\mathrm{density}_{it}$ and $v^{\text{avg}}_{it}$) apply to both highway and urban settings.
Traffic density is computed as the number of surrounding agents within an omni-directional perception region centered on the vehicle, reflecting general situational awareness. 
Average speed is computed using front and lateral perception zones, reflecting what the driver observes directly in the driving corridor. This distinction mirrors practical perception logic in modern sensing systems. 

Context determines profile activation through parameters $\beta_k \in \mathbb{R}^q$ for each  profile $k$. The activation weight is:

\begin{equation}
\pi_k(c_{it}) =
\frac{\exp(\beta_k^\top c_{it})}
{\sum_{j=1}^K \exp(\beta_j^\top c_{it})},\end{equation}

where $\pi_k(c_{it}) \in [0,1]$ and $\sum_{k=1}^K \pi_k(c_{it}) = 1$. If a contextual variable has a large positive coefficient in $\beta_k$, higher values of that variable increase the likelihood that profile $k$ governs behavior. Without this context step, profile weights would remain fixed over time, and the model would be unable to explain why the same driver behaves differently under different traffic or infrastructure conditions.

For each driver $i$ at time $t$, we define the time-varying behavioral state $\rho_i(t)$, which summarizes the driver’s current behavior in terms of $\Delta v$, $a$ and $h$ (through their RFF representation). At each timestep, we first form a predicted state

\begin{equation}
\rho_i^{\text{pred}}(t)
=
\mathord{\underbrace{(1-\alpha)\rho_i(t-1)}_{\text{persistence term}}} 
\hspace{0.7em} +
\mathord{\underbrace{\alpha \sum_{k=1}^K \pi_k(c_{it})\,\rho_k}_{\text{context-weighted mixture term}}},
\end{equation}

where $\alpha \in (0,1]$ controls how strongly context shifts the driver toward the population-level profiles. 
%The first term captures persistence (drivers do not change style instantly), while the second term captures context-induced adjustment.

The likelihood of the observed behavioral vector is:

\begin{equation}
p(x_{it} \mid \rho_i^{\text{pred}}(t))
=
\tilde{\phi}(x_{it})^\top
\rho_i^{\text{pred}}(t)
\tilde{\phi}(x_{it}),
\end{equation}

which measures how consistent the realized driving action $(\Delta v, a, h)$ is with the predicted state. After observing the actual behavioral vector $x_{it}$, the state is updated as:

\begin{equation}
\rho_i(t)
=
(1-\eta)\rho_i^{\text{pred}}(t)
+
\eta\,\tilde{\phi}(x_{it})\tilde{\phi}(x_{it})^\top,
\quad \eta \in [0,1].
\end{equation}

This update incorporates the observed relative speed, acceleration, and headway into the driver’s state representation. The parameter $\eta$ controls how strongly the current observation influences the state relative to its predicted value.

\subsection{Estimation and Interpretation}

The estimated parameters include the population-level profiles ${\rho_k}$, which encode interaction patterns; the context coefficients ${\beta_k}$, which determine how environmental conditions influence profile weights; and the scalars $\alpha$ and $\eta$, which govern contextual responsiveness and state adaptation. These are estimated by minimizing the negative log-likelihood:

\begin{equation}
\mathcal{L}
=
-\sum_{i,t}
\log p(x_{it} \mid \rho_i(t)),
\end{equation}

subject to the density-matrix constraints on $\rho_k$. Estimation used automatic differentiation (PyTorch), which computes exact gradients in a single forward-backward pass per optimization step at $O(1)$ cost, replacing coordinate-wise finite differences that require $O(P)$ forward passes where $P = \sum_k D^2 + Kq + 2$ is the total parameter count. For the configuration used here ($K{=}4$, $D{=}100$, $q{=}4$), $P > 40{,}000$, reducing per-epoch training time from approximately 3–-4 hours to under 15 minutes on the same hardware.

Eigen-decomposition of $\rho_k$ reveals dominant nonlinear behavioral modes and their relative importance. The temporal evolution of $\rho_i(t)$ provides insight into how drivers adapt behavior over time in response to context and experience.

\section{Simulation Analysis}

\subsection{Simulation Setup}

The proposed framework is evaluated using the Third Generation Simulation (TGSIM) trajectory datasets, which provide high-resolution vehicle trajectories sampled at 0.1\,s intervals. We use both the Foggy Bottom (urban intersection) \cite{tgsim_foggybottom_2024} and I-395 (freeway) \cite{tgsim_i395_2024} deployments to ensure coverage across heterogeneous traffic environments, spanning urban interactions and higher speed car-following dynamics. Raw trajectories were Gaussian-smoothed and filtered to retain only continuous trajectories. The final dataset contains 3,200,397 observations from 4,360 human-driven vehicle trajectories, of which 1,277 are from Foggy Bottom and 3083 from I-395. 

%Raw trajectories were Gaussian-smoothed to reduce sensor noise that can destabilize learning and distort state evolution. Records were further filtered to retain only continuous, non-interrupted trajectories, as temporal gaps or abrupt agent disappearance disrupt state propagation. This preprocessing ensures stable estimation and coherent behavioral evolution across time. The final dataset contains 3,200,397 observations from 4,360 human-driven vehicle trajectories, of which 1,277 are from Foggy Bottom and 3083 from I-395. 

The learned density-matrix profiles from the TGSIM data are evaluated in terms of spectral structure, context activation, behavioral characterization and geometric separation.

\subsection{Selection of Number of Behavioral Profiles}

The number of latent behavioral profiles $K$ is selected by comparing model fit and spectral structure across $K \in \{3,4,5\}$. For each candidate, multiple random initializations are trained under identical settings. The mean negative log-likelihood per observation $\left(\bar{\ell}\right)$ is recorded. Increasing $K$ from 3 to 4 reduces mean $\bar{\ell}$ from $0.658$ to $0.629$, indicating improved fit. Increasing to $K=5$ (mean $\bar{\ell}=0.629$) provides no improvement. Balancing likelihood, spectral richness, and parsimony, $K=4$ provides the most expressive and stable representation and is adopted for subsequent analyses.

\subsection{Spectral Structure of Behavioral Profiles}

Each behavioral profile $\rho_k$ represents a population-level driving regime. Within a profile, eigen-decomposition reveals one or more behavioral modes, where each mode corresponds to a dominant interaction pattern in the nonlinear RFF representation. 
Thus, profiles define regimes, while modes describe the internal structure of each regime.

Table~\ref{tab:eigenvalues} reports the leading eigenvalues of each learned density matrix $\rho_k$. Three of the four profiles converge to effectively rank-1 structures, indicating that these behavioral regimes are well characterized by a single dominant mode in the nonlinear feature space.

Profile~3 exhibits a qualitatively different structure, with two leading eigenvalues carrying approximately 71.5\% and 27.8\% of the spectral mass, and a third mode contributing 0.6\%. This multi-modal spectrum indicates a behavioral regime that cannot be reduced to a single pattern. Drivers activating this profile exhibit a probabilistic mixture of at least two distinct behavioral modes. The coexistence of rank-1 and mixed-rank profiles within a single model is a direct consequence of the density matrix parameterization, which permits each $\rho_k$ to learn its own effective dimensionality from the data without imposing a rank constraint.

\begin{table}[ht!]
\centering
\caption{Leading eigenvalues of learned behavioral profiles ($K{=}4$).}
\label{tab:eigenvalues}
\begin{tabularx}{\linewidth}{>{\centering\arraybackslash}X >{\centering\arraybackslash}X 
                              >{\centering\arraybackslash}X>{\centering\arraybackslash}X 
                              >{\centering\arraybackslash}X} 
\toprule
Profile & $\lambda_1$ & $\lambda_2$ & $\lambda_3$ & $\lambda_4$ \\
\midrule
1 & 0.9999 & 0.0000 & 0.0000 & 0.0000 \\
2 & 0.9999 & 0.0001 & 0.0000 & 0.0000 \\
3 & 0.7151 & 0.2783 & 0.0062 & 0.0003 \\
4 & 0.9999 & 0.0001 & 0.0000 & 0.0000 \\
\bottomrule
\end{tabularx}
\end{table}

\subsection{Context-Dependent Profile Activation}

\begin{figure}[ht!]
\centering
\includegraphics[width=0.7\linewidth]{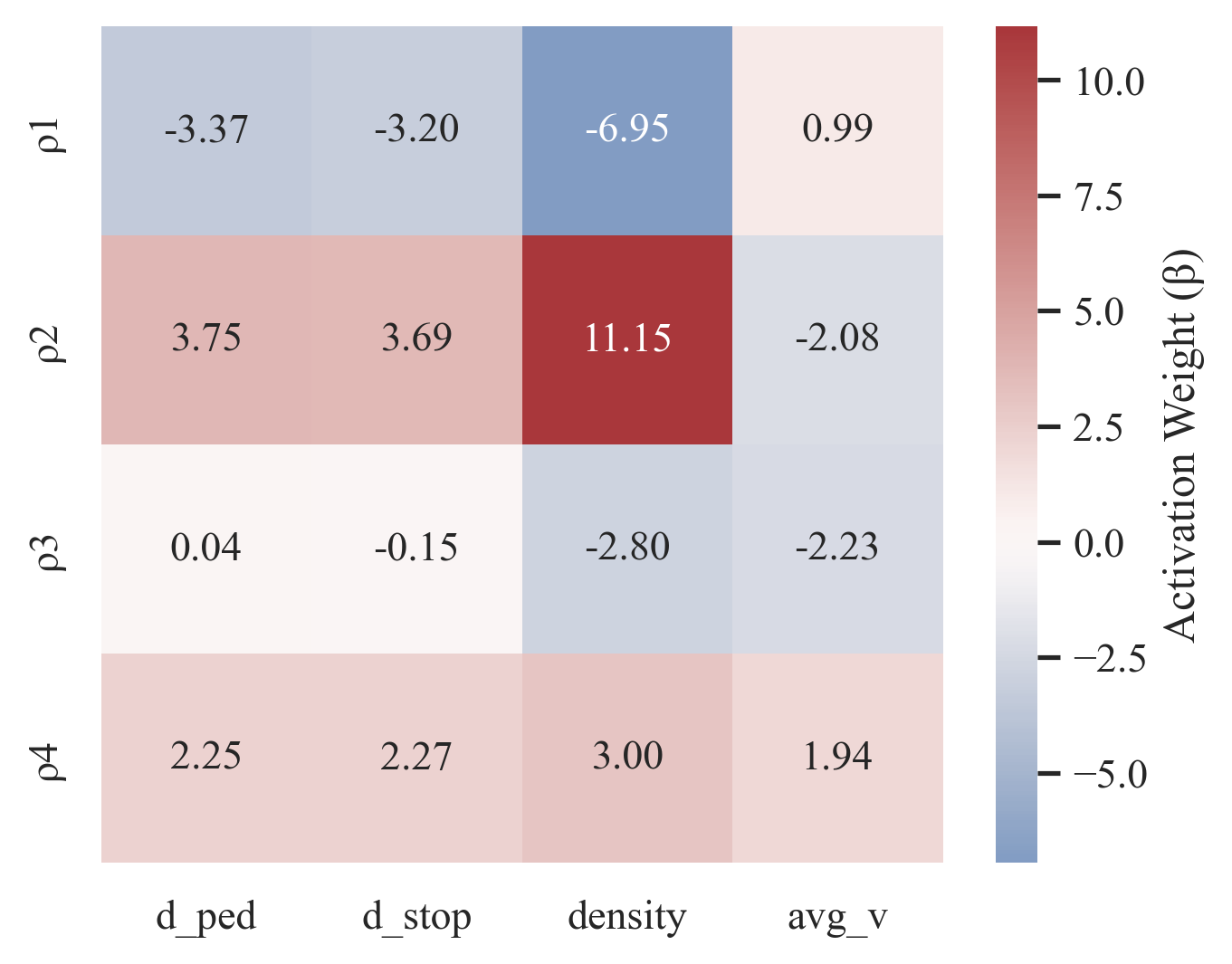}
\caption{Context activation coefficients ($\beta$) across identified behavioral profiles.}
\label{fig:beta}
\end{figure}

The learned activation weights $\beta_k$ govern activation weight of profiles, determining how environmental context modulates the influence of each behavioral profile on the driver state. The magnitude and sign of each coefficient (Figure~\ref{fig:beta}) indicate the contextual conditions under which a profile becomes dominant.

Profile~2 exhibits the strongest context sensitivity, with a large positive coefficient on traffic density ($\beta = 11.15$) and positive loadings on pedestrian and stop-sign proximity. This profile is preferentially activated in congested, interaction-rich environments where drivers must negotiate complex stimuli. Conversely, Profile~1 loads negatively on all spatial context variables, with its strongest suppression on density ($\beta = -6.95$), indicating activation in low-density, unconstrained driving conditions. 

Profile~4 shows moderate positive loadings across all four context variables, suggesting activation in environments with moderate traffic and higher average speeds. This is consistent with steady-state car-following on the freeway segment. Profile~3, the multi-modal profile, is relatively context-insensitive with near-zero loadings on pedestrian and stop-sign proximity but negative coefficients on density ($\beta = -2.80$) and average speed ($\beta = -2.23$). This weak contextual specificity is consistent with its mixed spectral structure: rather than being triggered by a single well-defined context, this regime captures transitional or heterogeneous driving states that span multiple environmental conditions.

Given the dataset combines urban intersection and freeway deployments, the emergence of both interaction-heavy and unconstrained patterns is expected. It also confirms that the activation mechanism depicts environment-specific behaviors without suppressing either setting. 

\subsection{Behavioral Characterization of Latent Profiles}

\begin{figure}[ht!]
\centering
\includegraphics[width=0.95\linewidth]{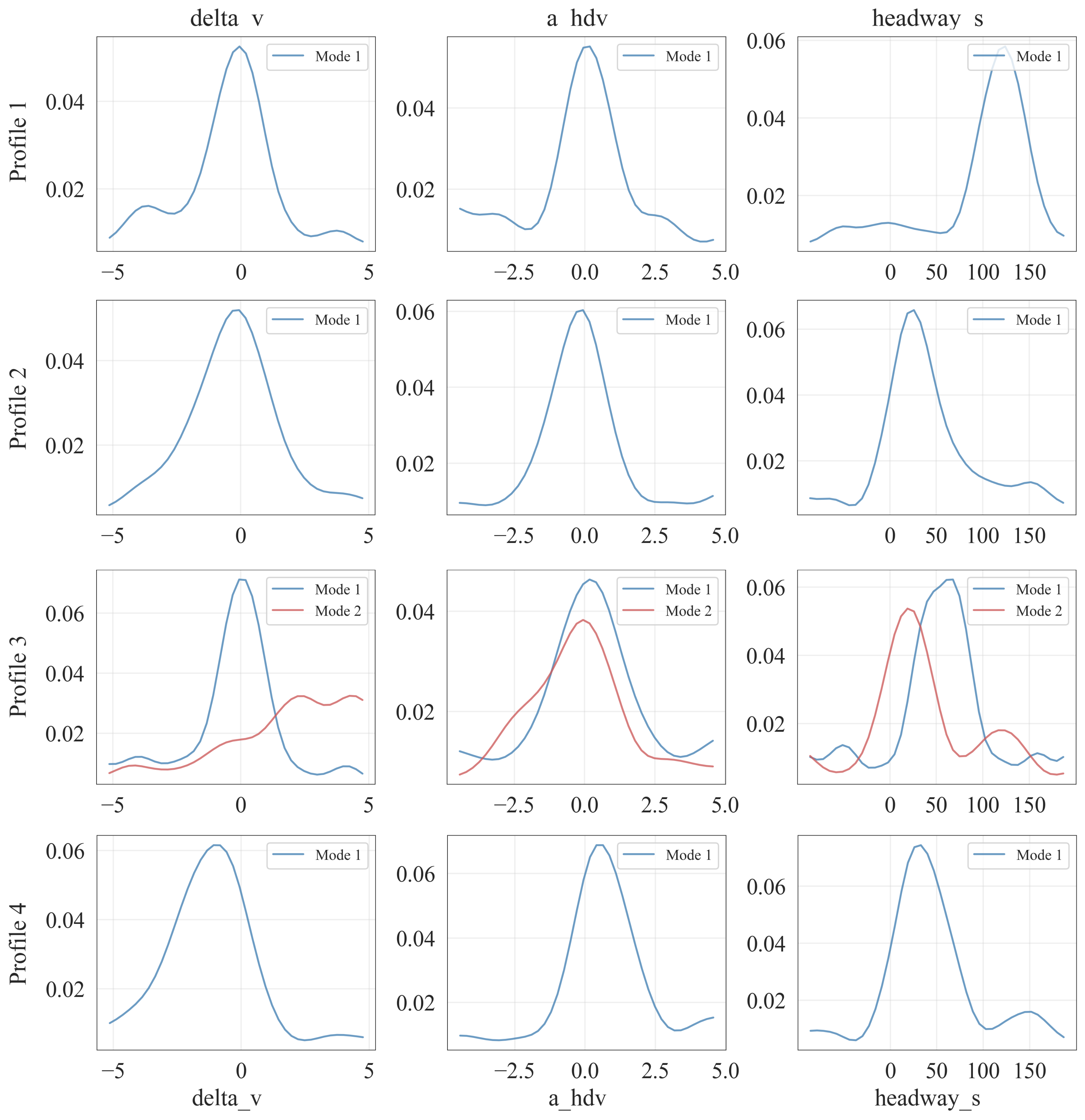}
\caption{Marginal activation of dominant eigenmodes across behavioral variables. Rows denote profiles and columns denote $\Delta v$, $a$, and $h$.}
\label{fig:marginal_activations}
\end{figure}

To interpret the learned profiles in behavioral terms, the top eigenvectors of each $\rho_k$ are projected back through the RFF mapping by evaluating $\tilde{\varphi}(x)^\top v_m$ over a dense grid in the original behavioral space, identifying the regions of $(\Delta v, a, h)$ where each eigenmode concentrates its activation. Figure~\ref{fig:marginal_activations} shows the marginal activation distributions in terms of behavioral variables for each behavioral profile.

The four latent profiles exhibit distinct longitudinal interaction patterns, aligning with their context activation coefficients.

\paragraph{Profile 1 (Free-Flow or Low-Constraint Regime)}  
As shown in Figure~\ref{fig:marginal_activations}, Profile 1 is characterized by the largest headways (mean $h \approx 83$\,m) and near-neutral acceleration, consistent with low-interaction or free-flow driving. Strong negative coefficients on pedestrian proximity, stop proximity, and density confirm activation when environmental constraints are minimal. The positive $v^{\text{avg}}$ coefficient further supports association with higher-speed, unconstrained conditions.

\paragraph{Profile 2 (Dense-Interaction Regime)}  
Profile 2 reflects mild deceleration or steady-state following with shorter headways (mean $h \approx 44$\,m) and minimal acceleration. Strong positive coefficients on pedestrian proximity, stop proximity, and density indicate that this regime is most active in dense, interaction-heavy environments. The negative $v^{\text{avg}}$ coefficient suggests predominance in lower-speed urban conditions.

\begin{figure}[ht!]
\centering
\includegraphics[width=0.58\linewidth]{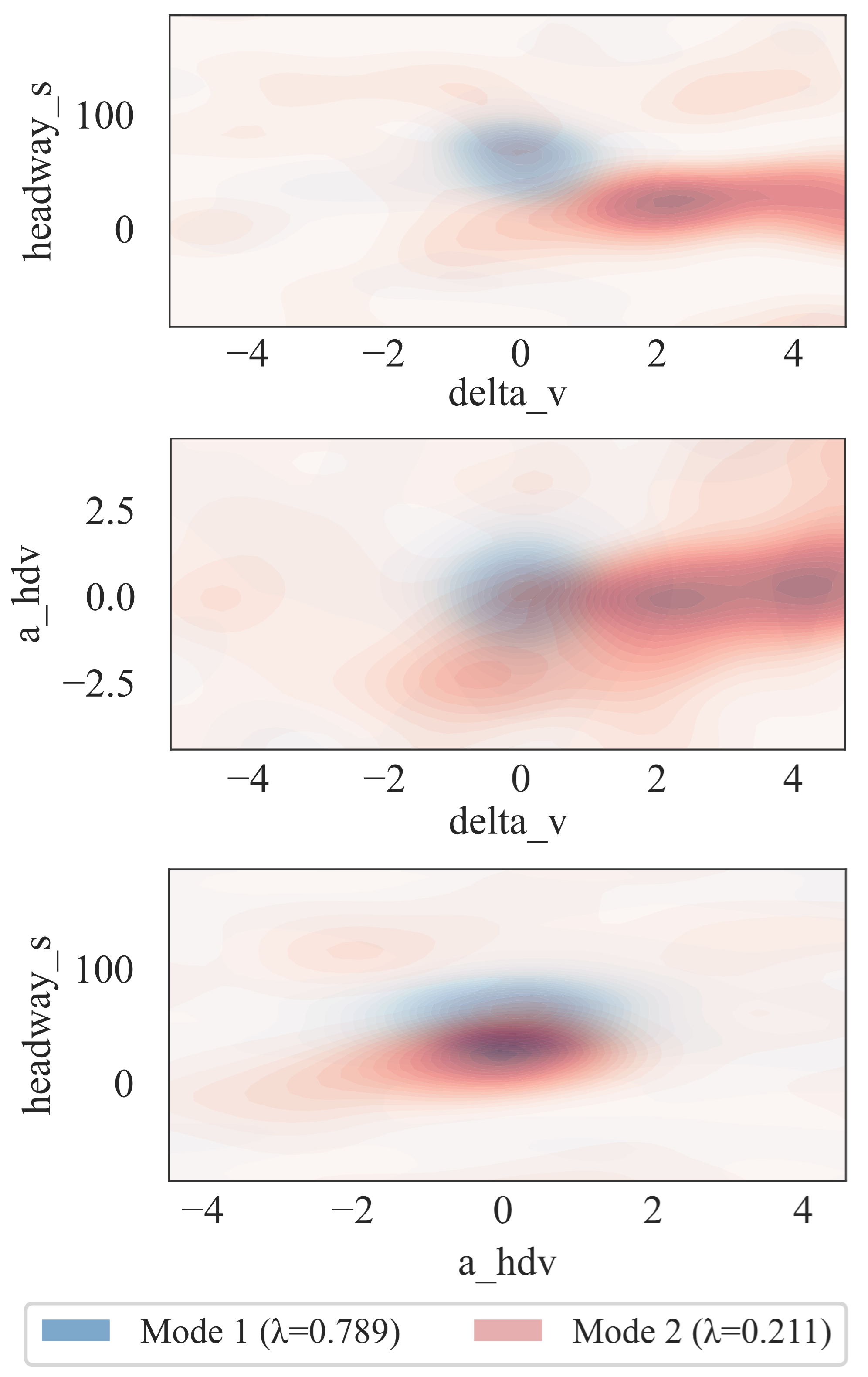}
\caption{Multimodal activation heatmaps for Profile 3. Shaded contours represent the localized activation of each mode, illustrating the structural diversity within the profile geometry.}
\label{fig:profile_3}
\end{figure}

\paragraph{Profile 3 (Multi-Modal Regime)}  
As shown in Table~\ref{tab:eigenvalues} and Figure~\ref{fig:profile_3}, Profile 3 exhibits two coexisting behavioral tendencies ($\lambda_1=0.79$ and $\lambda_2=0.21$). The dominant mode corresponds to moderate positive acceleration with relatively large headways (mean $h \approx 53$\,m), suggesting controlled progression under comfortable spacing. The secondary mode shifts toward higher relative speeds and shorter headways (mean $h \approx 41$\,m), reflecting a more assertive car-following response. The associated context coefficients show weak sensitivity to pedestrian and stop proximity but negative sensitivity to density and average speed, indicating activation in moderately constrained flow (e.g., midblock in an urban environment) rather than high-speed free-flow conditions.

\paragraph{Profile 4 (Responsive Following).}  
Profile 4 is characterized by negative relative speed (mean $\Delta v \approx -1.07$\,m/s) with positive acceleration and moderate headway. This pattern suggests drivers recovering speed while trailing a slower leader. Positive coefficients on $d_{\text{ped}}$, $d_{\text{stop}}$, and density indicate activation in environments with greater surrounding interaction, while the positive $avg_v$ coefficient suggests operation in generally faster traffic streams.

Overall, the learned profiles partition longitudinal behavior into (i) unconstrained free-flow operation (Profile 1), (ii) interaction-driven dense traffic behavior (Profile 2), (iii) a mixed transitional regime (Profile 3), and (iv) responsive speed recovery under moderate constraints (Profile 4). The alignment between behavioral statistics and context activation coefficients indicates coherent coupling between latent state structure and environmental conditions.

\subsection{Profile Geometry and Separation}

\begin{figure}[ht!]
\centering
\includegraphics[width=0.7\linewidth]{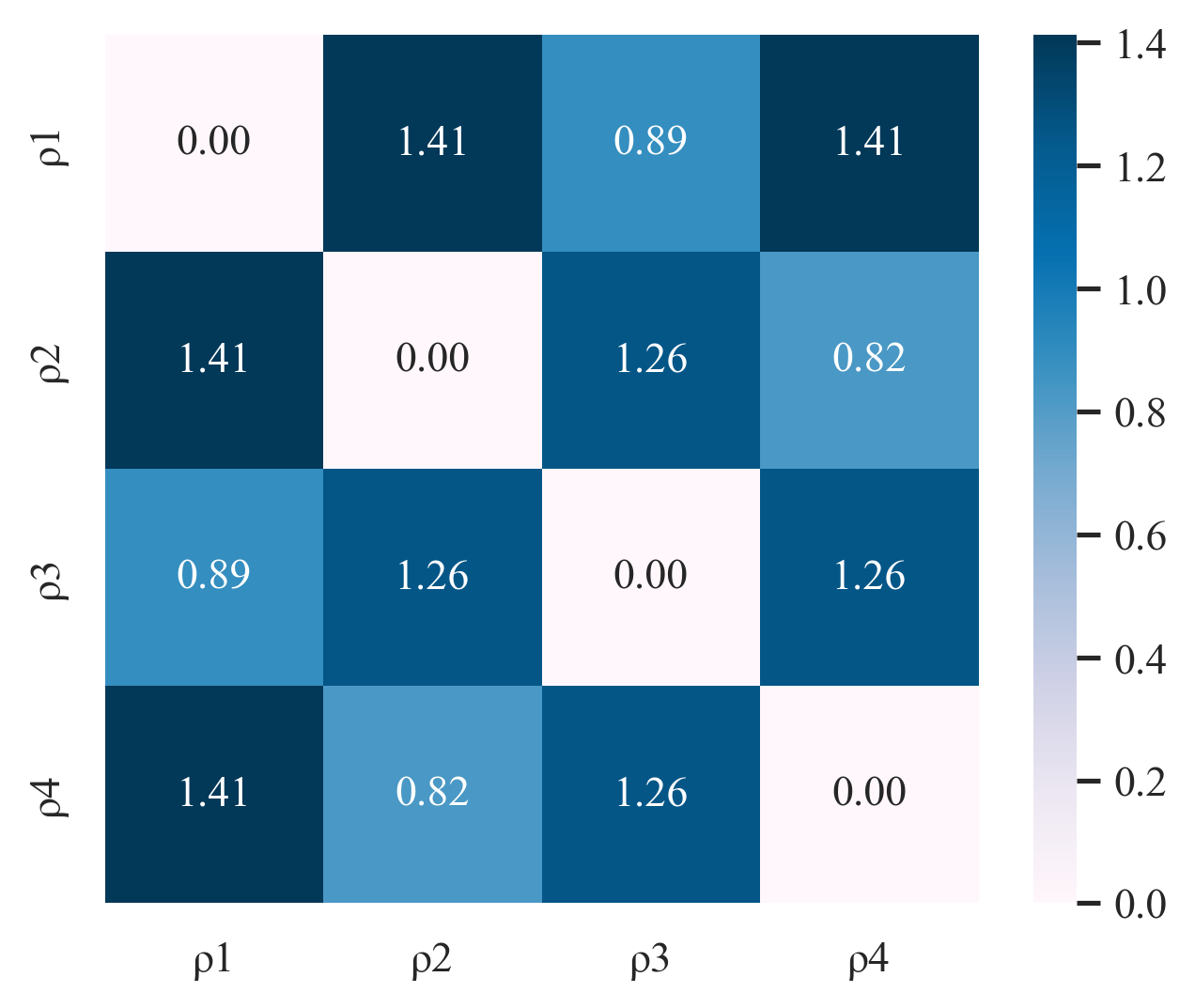}
\caption{Pairwise Frobenius distances between identified behavioral profiles. Numerical values represent the magnitude of divergence in profile geometry.}
\label{fig:geometry}
\end{figure}

Geometric separation between profiles is quantified using the Frobenius distance \(d_F(\rho_j, \rho_k) = \|\rho_j - \rho_k\|_F = \sqrt{\mathrm{Tr}[(\rho_j - \rho_k)^2]}\), which measures the total element-wise divergence between two density matrices. Figure~\ref{fig:geometry} reports pairwise distances. All off-diagonal values are well above zero, confirming that the four learned profiles occupy distinct regions of the nonlinear RFF representation. This separation is consistent with the behavioral interpretation, as the profiles correspond to clearly different interaction regimes (free-flow, dense interaction, multimodal, and responsive following).

Profile~3, the only multi-modal profile, is nearest to Profile~1 ($d_F = 0.893$). This proximity is consistent with its dominant mode, which shares relatively large headways and moderate acceleration with the low-constraint regime, while its secondary mode introduces an assertive interaction pattern.

The largest separations occur between Profiles~1 and~4 ($d_F = 1.412$) and between Profiles~1 and~2 ($d_F = 1.411$), reflecting the contrast between unconstrained free-flow behavior and interaction-heavy regimes.

\section{Conclusions}

This paper proposes a quantum inspired model of driver heterogeneity where each driver is represented by an evolving latent state encoded as a trace normalized positive semidefinite density matrix in a nonlinear feature space. Longitudinal observations based on relative speed acceleration and headway are embedded with normalized RFF so nonlinear structure can be captured while keeping the state compact. Population behavioral profiles are learned as density matrices and each driver state is updated through a persistence and context mixture prediction followed by an observation based correction using a quadratic likelihood consistent with the Born rule. Using high resolution TGSIM data from an urban intersection and a freeway the method identifies four parsimonious profiles with clear spectral structure including three near rank one regimes and one multimodal regime along with geometric separation in Frobenius distance. The context activation parameters align these profiles with traffic conditions such as density and proximity and the sequential updates make it possible to interpret how behavior shifts over time rather than assigning static driver types.

The approach is intentionally compact in its state and context activation, which aids interpretability, but it omits lateral behavior and richer interaction cues and depends on how context is defined. Future work should add lateral and multi vehicle features, learn more flexible but constrained context mappings, and use ablation tests to isolate context effects. Finally, broader comparisons to sequential baselines and transfer tests across sites and conditions would clarify when evolving states improve prediction of system level outcomes, including stability, capacity drop, and mixed autonomy effects.

\section*{ACKNOWLEDGMENTS}
This research was sponsored by NSF National Artificial Intelligence Research Resource Pilot (NAIRR).

%through award NAIRR250132. 

\bibliographystyle{IEEEtran}
\bibliography{references}

\end{document}